\documentclass[runningheads]{llncs}
\usepackage{graphicx}
\usepackage{cite}
\usepackage{hyperref}
\usepackage{footnote}
\usepackage{csquotes}
\usepackage{subfigure}
\usepackage{xurl}
\usepackage{graphicx}
\usepackage{csquotes}
\usepackage{amsmath,amssymb,amsfonts}
\usepackage{algorithm}
\usepackage{framed,multirow}

\usepackage{mathrsfs}
\usepackage{float}
\usepackage{hyperref}
\usepackage[T1]{fontenc}

\begin{document}
\title{Machine learning equipped web based disease prediction and recommender system}
\titlerunning{Web based disease prediction and recommender system}

\author{Harish Rajora\orcidID{0000-0002-7171-0064} 
\and Narinder Singh Punn\orcidID{0000-0003-1175-1865}
\and Sanjay Kumar Sonbhadra\orcidID{0000-0002-7457-9655}\and Sonali Agarwal\orcidID{0000-0001-9083-5033}}
 \authorrunning{H. Rajora et al.}
%
\institute{Indian Institute of Information Technology, Allahabad, India \\
\email{\{mit2019018, pse2017002, rsi2017502, sonali\}@iiita.ac.in}}

\maketitle              
\begin{abstract}
Worldwide, several cases go undiagnosed due to poor healthcare support in remote areas. In this context, a centralized system is needed for effective monitoring and analysis of the medical records. A web-based patient diagnostic system is a central platform to store the medical history and predict the possible disease based on the current symptoms experienced by a patient to ensure faster and accurate diagnosis. Early disease prediction can help the users determine the severity of the disease and take quick action. The proposed web-based disease prediction system utilizes machine learning based classification techniques on a data set acquired from the National Centre of Disease Control (NCDC). $K$-nearest neighbor ($K$-NN), random forest and naive bayes classification approaches are utilized and an ensemble voting algorithm is also proposed where each classifier is assigned weights dynamically based on the prediction confidence. The proposed system is also equipped with a recommendation scheme to recommend the type of tests based on the existing symptoms of the patient, so that necessary precautions can be taken. A centralized database ensures that the medical data is preserved and there is transparency in the system. The tampering into the system is prevented by giving the no \textquote{updation} rights once the diagnosis is created.

\keywords{Disease prediction, Healthcare, Web, Ensemble learning}
\end{abstract}
\section{Introduction}
Organization of the medical data is always a challenging task for betterment of modern healthcare system. It is evident that mostly the medical data is exclusive only for the healthcare organizations which is only practised by elite hospitals. Following this, the idea of the proposed platform is to offer global reachability of the medical data that is not restricted just to one hospital and its subsidiaries, but to local hospitals and clinics (if the doctor is registered), while also securing the patient's identity. This platform can be used as an excellent source for the collection of data of the common symptoms and the underlying disease. The data can be used for various research activities and to generate new real-life data sets. Insurance companies can also make use of authentic data to bring down the fraudulent claims which on an average ranges between $\sim$10 to $\sim$15 \cite{insurancefraud}. The system is equipped with a prediction algorithm that works in the back-end to predict the disease from the given symptoms.

With immense applications of deep learning and machine learning across various domains and applications such as classification, detection, recognition, localization, etc. \cite{punn2019crowd, punn2020chs, sonbhadra2020target, zhang2019empirical}. These techniques have been used for several application domains such as the future box office movie collections, price of a house with given parameters or bankruptcy prediction, human face recognition systems and voice recognition, etc. \cite{BarbozaaAltman,MuHassoun}. The ensemble approach \cite{PaulIba,BashirKhan} has been proved to be very effective and more accurate \cite{BashirKhan2} in many medical applications such as heart disease prediction \cite{HazraMukherjee,PowarGhorpade,ChaitraliApte}, cancer prediction \cite{PaulIba}, hepatitis disease prediction \cite{DuttaBandyopadhyay}, diabetes prediction \cite{PathirageSilva} and Parkinson’s disease prediction \cite{RaoVital}. With this motivation, to predict plausible disease an ensemble approach is proposed in the present research by combining the features of three state-of-the-art classification algorithms:  naive bayesian, $K$-NN and random forest. The selection of these algorithms follows from various research papers \cite{MuHassoun,DuttaBandyopadhyay,PatilSagar} which are focused on determining how the different classification algorithms behave with respect to the different data chunks and different applications. Following this, the ensemble approach is combined with the voting mechanism \cite{PaulIba,DuttaBandyopadhyay,RamasamyNirmala} (discussed in the sub-sections \ref{4.2}) to provide the best prediction of the disease.

The main contribution of this paper is as follows:
\begin{itemize}
    \item \textit{Disease prediction}: This paper provides an algorithm to predict the disease based on the given symptoms using state-of-the-art machine learning classification algorithms.
    \item \textit{Analyze ensemble results}: Combine the results of the individual algorithms with voting mechanism discussed in sub-section \ref{4.2}.
    \item \textit{Web-based interactive platform}: The research also presents an interactive platform (Raahat\footnote{\url{https://bdalabiiita.pythonanywhere.com/}\label{Raahat}}) integrated with the disease prediction system along with other important features for a more robust healthcare management system. 
    
\end{itemize}

The paper is divided into six major sections. Section 2 describes the motivation behind the project. Section 3 explains the background knowledge required to understand the various algorithms used. Section 4 is a brief introduction to the components of the platform including prediction and voting mechanism. Section 5 contains the results and output of the algorithm. Finally, the concluding remarks are presented in Section 6.

\section{Motivation}
A more robust and transparent healthcare infrastructure is the primary global need of most of the countries. The non-existence of such a system causes delayed diagnosis, transfer of misinformation and lack of information among the doctors and people. Healthcare management systems have proved to be beneficial in some developed countries such as the UK and USA, where a patient’s complete medical record is accessible to every registered doctor for better and quicker diagnosis. The system can also help if the patient is unconscious and has no alibis. Hence, a centralized system helps to minimise the chances of wrong medication and other such mishaps to ensure better recovery chances.

The observations and efforts to cover the gaps in the system have also paved the way to write a disease prediction algorithm and integrate it with the application. The motivation comes from the fact that people try to diagnose themselves by their previous knowledge which is not a right way, thereby risking the severity of the disease. For example, a running nose is instantly attributed to cold, stomach ache to overeating/acidity etc. While they might be correct in a small fraction of the cases, the people are unaware of the consequences of the disease they are suffering from. The diagnosis at the later stage might result in increased expenditure and higher complexities which are difficult to treat. A disease prediction module in the system that can help to diagnose sitting at home and take actions quickly in case the disease is serious. The disease prediction can also be used for confirmation of the prediction from the doctor and optimizing it frequently for better results.

\section{Literature review}
Deep learning and machine learning technologies are playing active role in various sectors to develop intelligent applications~\cite{punn2020multi, agarwal2020unleashing}. A disease prediction system is an important aspect of a developing nation as part of strong healthcare infrastructure. With this notion, the present research proposes a machine learning equipped web-based disease prediction system along with a recommender system to ensure a more strategic and robust healthcare environment. With deep insight into the existing e-healthcare system, it is evident that there are few disease specific solutions are available; but a diverse disease prediction system considering all common diseases is still missing (to the best of our knowledge). 

Kumar et al. \cite{PaulIba} has focused on the prediction of the cancer while Bashir et al. \cite{BashirKhan} proposed a solution to predict the heart disease for a patient. In similar approach, Dutta et al. \cite{DuttaBandyopadhyay} proposed an approach for hepatitis disease prediction, Rao et al. developed a parkinson's prediction approach \cite{RaoVital} and Pathirage et al. \cite{PathirageSilva} offered a diabetic prediction system. All these methods are algorithmic approaches and do not offer user interfaces for end users. The absence of a GUI creates a gap between the system and the user. Following this, the present research focuses on the common disease symptom (including rare) to cater to the needs of a large group of people and offers a user friendly web application. For disease prediction the present research uses the classification models: $K$-NN, naive bayes and random forest, and fuses the results using an ensemble approach with a novel voting scheme to ensure accurate prediction (covered in following subsection).

\subsection{Classification strategies}
The following section explains the background knowledge required to understand the proposed methodology. This section describes the classification algorithms used in the voting mechanism.

\subsubsection{$K$-NN}
The $K$-NN \cite{MudaliarGarg} algorithm classifies the new data point (the one that needs to be classified) based on the most similar classes that are known as neighbors ($N$). In this research, the data set is run multiple times for different values of $N\in[1,15]$ and the most promising value of $N$ is selected for further experiments, where the distance is determined by the Euclidean distance as shown in Eq. \ref{eq1}:
\begin{equation}
    Dist(X^n , X^m) = \sqrt{\sum_{i = 1} ^ {D} (X_{i}^n - X_{i}^m)^2}
    \label{eq1}
\end{equation}

\subsubsection{Random forest}
The random forest \cite{BarbozaaAltman} algorithm takes decision trees as the basic block in the construction of the model. The idea behind the implementation of random forest is that taking output from one decision tree may not be entirely reliable. Since there is a lot of variance involved in generating the final output, it may not necessarily be closest to the actual answer. Hence, random forest creates multiple decision trees and analyze them together. Random forest also has an edge over decision trees as decision trees are prone to overfitting. Random forest selects random smaller samples and builds tree from them overcoming the disadvantage of overfitting in decision trees. Therefore,  multiple decision trees approach used in random forest generates a better output than a single decision tree.

\subsubsection{Naive bayes}
The naive bayes classification \cite{MudaliarGarg} technique works on the Bayes' theorem popularly used in probability analysis in mathematics and computer science. Naive bayes algorithm considers each property as an independent property contributing to the final classification. Due to this reason, Naive Bayes has been acclaimed of its accuracy and reliability across various researches.
The Bayes' theorem used in the Naive Bayes algorithm can be represented using Eq. \ref{eq2} and Eq. \ref{eq7}:
\begin{equation}
    P(c|x) = \frac{P(x|c)P(c)}{P(x)}
    \label{eq2}
\end{equation}

\begin{equation}
    P(c|X) = P(x_{1}|c) * P(x_{2} |c) * .... * P(X_{n} | c) * P(c)
    \label{eq7}
\end{equation}

\subsubsection{Ensemble}
The voting algorithm has also been used in various researches to improve the overall efficiency of the model. A voting algorithm uses multiple machine learning algorithms and a method to leverage the output from these individual algorithms. Voting algorithms are a better way of devising prediction based approaches since more than one model is used for providing reliable and efficient results. Wide variety of voting approaches have been proposed, for instance, Dutta et al. \cite{DuttaBandyopadhyay} have used the majority voting technique in their research. In the majority voting technique, the votes are counted for each prediction given by the algorithms. Therefore, if a prediction \textquote{X} is predicted by the majority of the algorithms, then \textquote{X} disease is the predicted disease. Xiaoyan et. al \cite{MuHassoun} has presented two voting strategies; majority and plurality, where in majority voting the final classification is declared based on half of the votes, while plurality voting selects the classifier having the highest number of votes.

The present research takes the confidence score of the three classification algorithms namely $K$-NN, random forest and naive bayes. The voting algorithm then creates another model taking the input as the individual models. In this paper, weighted voting algorithm is used  \cite{MuHassoun,BashirKhan2} where a weight parameter is defined to prioritize a classifier. The \textquote{weight} used for individual algorithms is the mean accuracy of the model. A model with higher mean accuracy shows a better performance compared to other models.

\section {Proposed methodology}
The proposed solution covers two major phases: 1) Web application development and 2) disease prediction via machine learning. The web application is for the user to interact with the system while the disease prediction system is used to predict the disease based on the given symptoms.

The prediction system is based on the ensemble classifier technique using naive bayes, $K$-NN and random forest. The back-end has been developed in the python programming language with the support of Django framework. The front-end and back-end are connected with the Jinja2 engine and the database used is SQLite. The module completely comprises of a GUI for the user to interact and a back-end server with integrated machine learning system for disease prediction. The following section presents the detailed description of the overall system.

\subsection{Web application}

The web application is developed for the module so that the user can interact with the platform. The web application consists of two major parts:
\begin{itemize}
    \item Front-End (majorly developed in JavaScript and HTML)
    \item Back-End (majorly developed in Python)
\end{itemize}

The front-end offers a GUI with minimal complexities, whereas the back-end of the website is developed that enables the communication between the front-end and the back-end using Jinja2 engine. The disease prediction system uses asynchronous calls from front-end to back-end to generate the output by taking the symptoms from the user without generating a new request.

The web application offers the following functionalities:
\begin{itemize}
    \item \textbf{Login and registration system} - offers sign-up and sign-in facilities into the system.
    
    \item \textbf{History lookup} - helps in fetching the history of a patient by the doctors or patients.
    
    \item \textbf{Creating medical log} - This is only available for doctor user. With the help of this module, the doctors can create the medical entry for a patient that will be stored into the database.
    
    \item \textbf{Predicting the disease based on symptoms} - helps people predict their diseases by selecting the symptoms from the list given on the platform.
    
    \item \textbf{An informative section for basic health related schemes} - provides a scheme structure to manage scheme beneficiaries. 
    
    \item \textbf{Quick diagnosis for the over-the-counter medicines} - helps the people in getting OTC medicines along with the predicted disease.
\end{itemize}

\begin{figure}
\includegraphics[width= \textwidth]{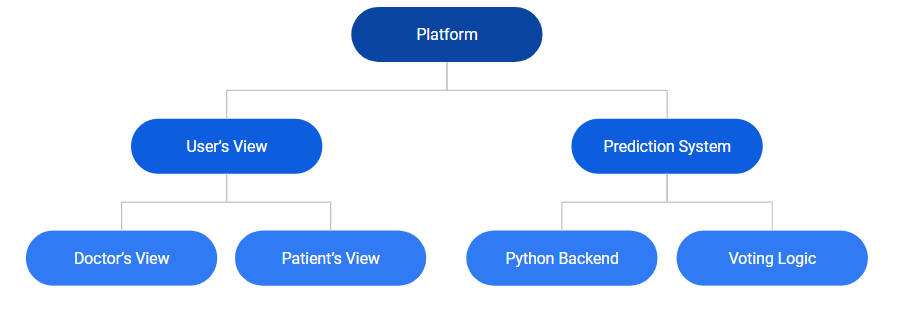}
\caption{High level schematic representation of proposed framework.}
\label{fig:schema}
\end{figure}

Figure \ref{fig:schema} shows the modules and functionalities described above with a high level schematic representation. The platform is divided into two major segments: one is the user's view and another is the disease prediction system. Since there are different rights of a doctor and a patient, the user's view is further divided into two more views. The prediction system comprises the python back-end which asynchronously communicates with symptoms and the voting logic to generate the output.

\subsection{Disease prediction system}

The prediction system is based on the ensemble classifier techniques using naive bayes, $K$-NN and random forest. To optimize the prediction techniques, a voting mechanism is used for assigning values to various classifiers which would act as weights for the system. The higher the weight, the more priority will be given to the classifier’s output. The voting is done based on the mean accuracy score of the classifier which is described in the next section. The final result can then be conveyed to the user through the model and controller of the Django framework.

\subsubsection{Voting mechanism}
\label{4.2}
This section focuses on the voting mechanism of the disease prediction system used in the application as a module. The voting classifier takes into account the machine learning models used to fit the data set and the weights assigned to each classifier. 

The voting mechanism used in the module takes into account the mean accuracy score of the classifiers. First, each classifier is run and a mean accuracy score is generated providing the confidence of the classifier on the result. The same mean accuracy score is then used as weight for the voting mechanism and the result along with the accuracy is generated. The prediction results are provided to the user along with the accuracy so that the transparency is maintained in the system. Such a voting system is more effective instead of using any one of the classifiers as standalone for disease prediction.

\section{Results}
The experiments described in this paper are done on the data set taken from National Centre for Disease Control. The data set contains the data surveyed exhaustively with the most occurring symptoms in the patients. These symptoms are taken as final entries to construct the data set. There are a total of 4921 such unique entries in the data set. A single entry may contain a similar identified disease such as fungal infection but the symptoms in two entries for the same disease are different. This data set is then organised into another data set by processing the raw entries.

The new modelled data set contain symptoms as column names and the rows denote the identified disease. For every symptom occurring for a disease, the column entry is marked as 1 for that symptom while other entries are marked as 0. The data set is then split into the test data and train data with a ratio of 20:80. Each individual algorithm is run to fit them on the training data set and the experimental results are discussed in this section. The benchmark performance evaluation matrices used are presented in Eq. \ref{eq3}, Eq. \ref{eq4} and Eq. \ref{eq5}

\begin{equation}
    Accuracy = \frac{TP + TN}{TP +TN +FP +FN}
    \label{eq3}
\end{equation}

\begin{equation}
    Precision = \frac{TP}{TP + FP}
    \label{eq4}
\end{equation}
    
\begin{equation}
    Recall = \frac{TP}{TP + FN}
    \label{eq5}
\end{equation}    

The accuracy defined in Eq. \ref{eq3} is the measure of correct predictions done by the models. The precision defined in Eq. \ref{eq4} signifies the positive instances are classified as positive in the data set. The recall is the measure of correctly identifying true positives (Eq. \ref{eq5}).
The disease prediction algorithm is run on $K$-NN, random forest and naive bayes individually before combining them in the ensemble approach as described in the output section. 

\begin{figure}
\includegraphics[scale=0.6]{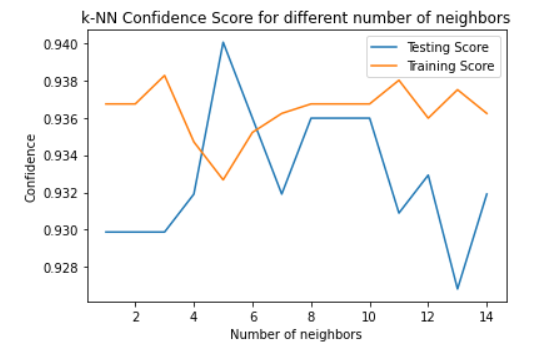}
\centering
\caption{The figure demonstrates the performance of different neighbours (1 to 15) over the data-set matched along the confidence score and number of neighbors axes.}
\label{fig:knngraph}
\end{figure}

For the $K$-NN classifier, 1 to 15 neighbours have been experimented and a result is generated to bring the maximum confidence score. To find the optimal value of the neighbours, each value of $N$ has been experimented for $\alpha$ times, where $\alpha=50$. Figure \ref{fig:knngraph} demonstrates the performance of $K$-NN with different neighbors.

\subsection{Experimental analysis}
\begin{figure}[h!]
    \includegraphics[width=\textwidth]{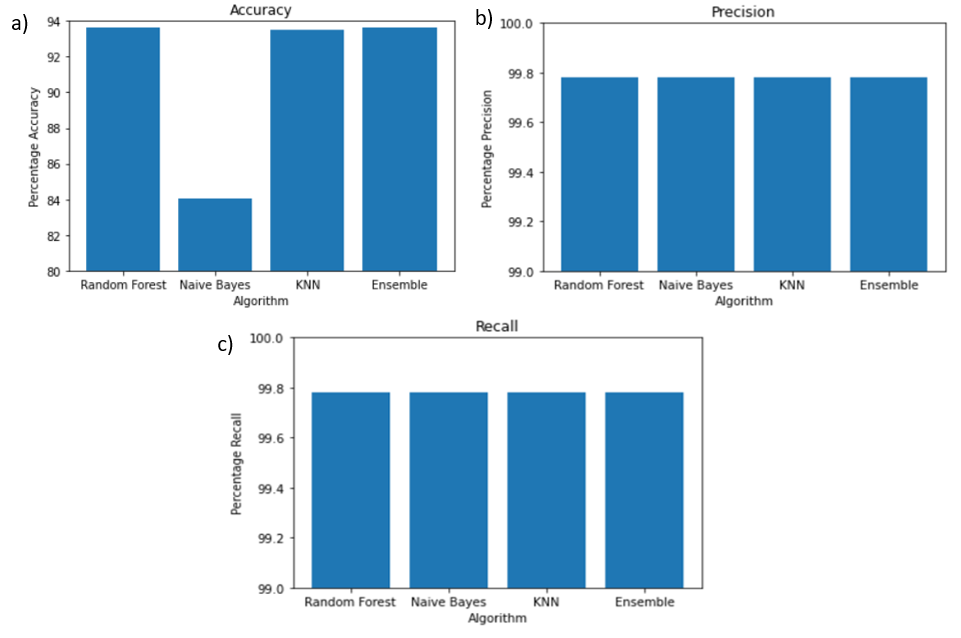}
    \caption{(a)Accuracy Score (b)Precision Score (c)Recall Score}
    \label{fig:scores}
\end{figure}

The disease prediction algorithms take a set of symptoms from the user as input. On the platform, the symptoms can be selected from the drop-down menu and the algorithms analyze them in the back-end. The symptoms selected by the user are then analyzed by each of the algorithms providing the output as the mean accuracy. The output is then used as the weights for those algorithms in the ensemble classifier. In the present section, the experimental symptoms considered are swelled lymph nodes, phlegm, redness of eyes, typhos, unsteadiness, enlarged thyroid. The confidence score values generated by the algorithms is shown in the Fig. \ref{fig:scores} and Table \ref{table:tab1}. 

\begin{table}
\centering
\caption{Confidence score values of classifiers}
\begin{tabular}{ |c|c|c|c| } 
\hline
\textbf{Classifiers} & \textbf{Accuracy} & \textbf{Precision} & \textbf{Recall}\\
\hline
Random Forest & 93.65 & 99.7818 & 99.7818\\
\hline
Naive Bayes & 84.02 & 99.7818 & 99.7818\\
\hline
$K$-NN & 93.53 & 99.7823 & 99.7823\\
\hline
Ensemble & 93.65 & 99.7818 & 99.7818\\
\hline
\end{tabular}
\label{table:tab1}
\end{table}

\begin{figure}
\begin{center}
    \includegraphics[scale = 0.5]{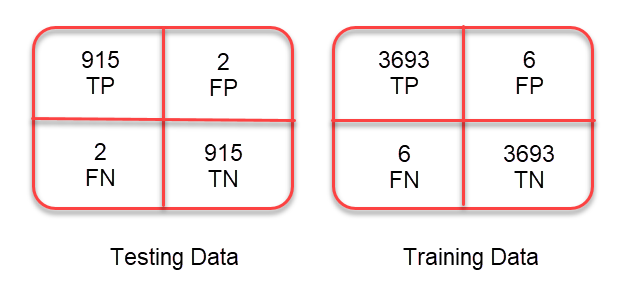}
    \caption{Confusion matrix testing data}
    \label{fig:confusion_matrix}
\end{center}
\end{figure}

For the experimental symptoms taken in this study, the naive bayes algorithm generated the least score. The score generated by naive bayes is 84.02, while random forest generated 93.65 and $K$-NN generated a score of 93.53. These numbers will be taken as weights in the ensemble model and the prediction will be run again with defined votes. In this experiment, the ensemble model generated a score of 93.65 which is the highest score and adapted to the best model among the three algorithms. The true positives, false positives, true negatives and false negatives calculated by the model is shown as a confusion matrix in the Fig.  \ref{fig:confusion_matrix}.

\begin{figure}
\centering
    \includegraphics[width=0.8\textwidth]{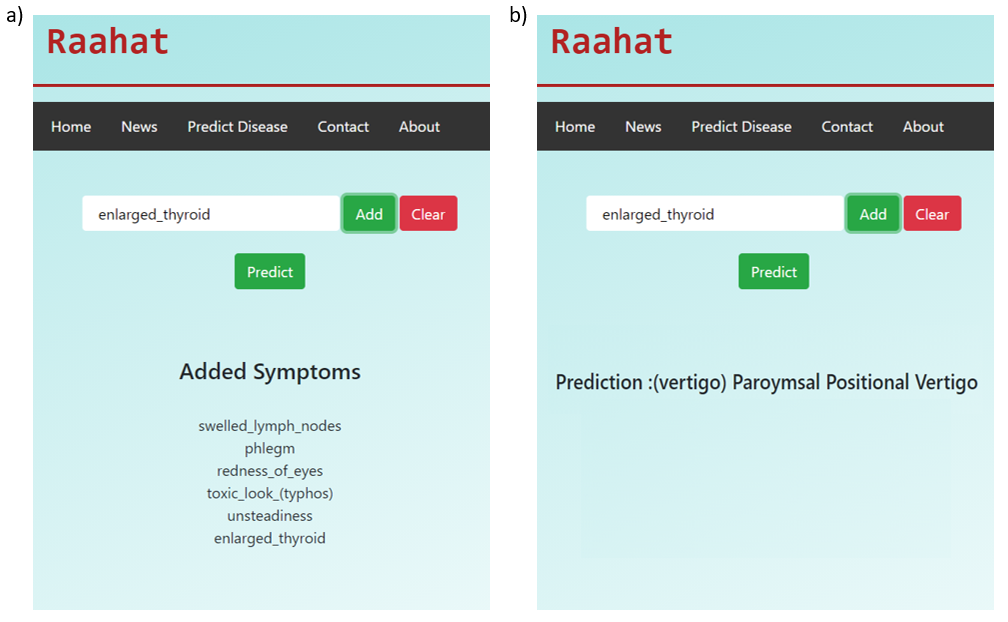}
    \caption{Raahat portal: (a)Adding symptoms on the portal (b) Output as predicted disease}
    \label{fig:symptoms_added}
\end{figure}

The platform boasts a front-end which takes on the input and present the output after processing it through the algorithm described in the paper. Once the symptoms are added, the user can press the \enquote{predict} button to pass the symptoms through AJAX call to the back-end. The predicted disease is then shown in the output. The screenshot after adding the symptoms in the application along with the predicted disease is shown in Fig. \ref{fig:symptoms_added}.


\section{Conclusion}
A disease prediction system is a helpful tool in uplifting the healthcare system of the country. A disease prediction system not only helps in \textquote{predicting a disease} but also in curating medical data, enhance research activities and control fraudulent activities. The present research offers a machine learning equipped web-based prediction system along with a recommendation system for a wide range of users. The ensemble approach is used as machine learning model that uses a weighted voting mechanism to prioritize better-performing algorithms on a given set of symptoms. The ensemble used in this paper involves three individual algorithms namely $K$-NN, naive bayes and random forest. Ensemble approaches prove to be more effective since the model uses more than one algorithms for an efficient and reliable result. The experiment is performed using a data set acquired from NCDC which presents the symptoms for a corresponding disease. A web-based application is also developed to help people interact and make use of the disease prediction system with a GUI. The experimental results show that the ensemble approach is adaptive to the best model and with the symptoms considered for experiments the ensemble approach adapted to the random forest.

\section*{Acknowledgment}
This research is supported by “ASEAN- India Science \& Technology Development Fund (AISTDF)”, SERB, Sanction letter no. – IMRC/AISTDF/R\&D/P-6/2017. Authors are also thankful to the authorities of “Indian Institute of Information Technology, Allahabad at Prayagraj”, for providing us with the infrastructure and necessary support.
\bibliography{references}
\bibliographystyle{splncs04}

\end{document}